\documentclass{article}
\usepackage{spconf,amsmath,graphicx}
\usepackage{amsfonts}
\usepackage{epsfig,amssymb,latexsym,graphics,float}
\usepackage{setspace,subfig}
\usepackage{citesort}
\usepackage{amsopn}
\usepackage{amssymb,times}

\title{Robust video object tracking using particle filter with likelihood based feature fusion and adaptive template updating}
\name{Yi~Dai$^{\dag}$ \qquad Bin~Liu$^{\star,\dag,\ddag}$
\thanks{$^\star$ Address correspondence to bins@ieee.org. This work
was partly supported by the National Natural Science Foundation
(NSF) of China under grant Nos. 61302158 and 61571238, the NSF of Jiangsu province
under grant No. BK20130869 and the Natural Science Research
Project of Jiangsu Province Department of Education under
grant No. 13KJB520019.}}
\address{$^{\dag}$ School of Computer Science and Technology, Nanjing University of Posts\\
and Telecommunications, P.R. China\\ 
$^{\ddag}$ State Key Laboratory for Novel Software Technology, Nanjing University, P.R. China}

\begin{document}
%
\maketitle
\begin{abstract}
A robust algorithm solution is proposed for tracking an object in complex video scenes. In this solution, the bootstrap particle filter (PF) is initialized by an object detector, which models the time-evolving background of the video signal by an adaptive Gaussian mixture. The motion of the object is expressed by a Markov model, which defines the state transition prior. The color and texture features are used to represent the object, and a marginal likelihood based feature fusion approach is proposed. A corresponding object template model updating procedure is developed to account for possible scale changes of the object in the tracking process. Experimental results show that our algorithm beats several existing alternatives in tackling challenging scenarios in video tracking tasks.
\end{abstract}
\section{Introduction}\label{sec:intro}
Object tracking in videos is an important task in many computer vision applications such as surveillance, augmented reality, human-computer interfaces, medical imaging and so on.
A straightforward strategy is to detect the target and determine its position frame by frame \cite{stauffer1999adaptive}. This process ignores the temporal correlation in the sequence of frame images and thus is incapable of dealing with occlusion cases. An alternative strategy is to use the state space to model the underlying dynamics of the tracking system. In this context, the particle filter (PF), also known as sequential Monte Carlo, is the most popular approach as it is neither limited to linear systems nor requires the noise to be Gaussian \cite{arulampalam2002tutorial,doucet2000sequential}.

In practice, prior to running a PF tracker, an object detector module is required to determine the object's initial position and an object template model based on a specific feature \cite{takala2007multi}. Therefore the performance of a PF tracker largely depends on the quality of the detector module and the selected features under use. The commonly used features include such as color \cite{nummiaro2003adaptive,yang2005fast}, edges \cite{vacchetti2004combining,yilmaz2006object,yang2005fast}, texture \cite{vacchetti2004combining}, motion\cite{takala2007multi} and so on.

Each tracker algorithm is successful with respect to a particular type of scene, e.g., the PF tracker using color feature is robust against noise and partial occlusion, but suffers from illumination changes, or presence of confusing colors in the background. In addition, the robustness of such methods is usually dependent on an invariant object template model. Hence the tracking performance may degrade rapidly due to scale changes or rotations of the object in the tracking process. 

Motivated by the challenges in object tracking in complex scenarios, we propose a novel PF solution, which combines an adaptive Gaussian mixture (GM) based object detector and makes use of the color and texture features of the object extensively in a theoretically sound manner. To account for possible scale changes of the object, an object template model updating procedure is also developed and embedded in this algorithm. The contribution of this work includes a likelihood based feature fusion method in the context of video object tracking, the corresponding online method to update the object template model, and the original mixture of the above components with PF and the GM based detector, which together yield a reliable and fast performance for object tracking in complex video scenes.

\section{The proposed approach}
\begin{figure}[!htb]
\begin{tabular}{c}
\centerline{\includegraphics[width=3in,height=5in]{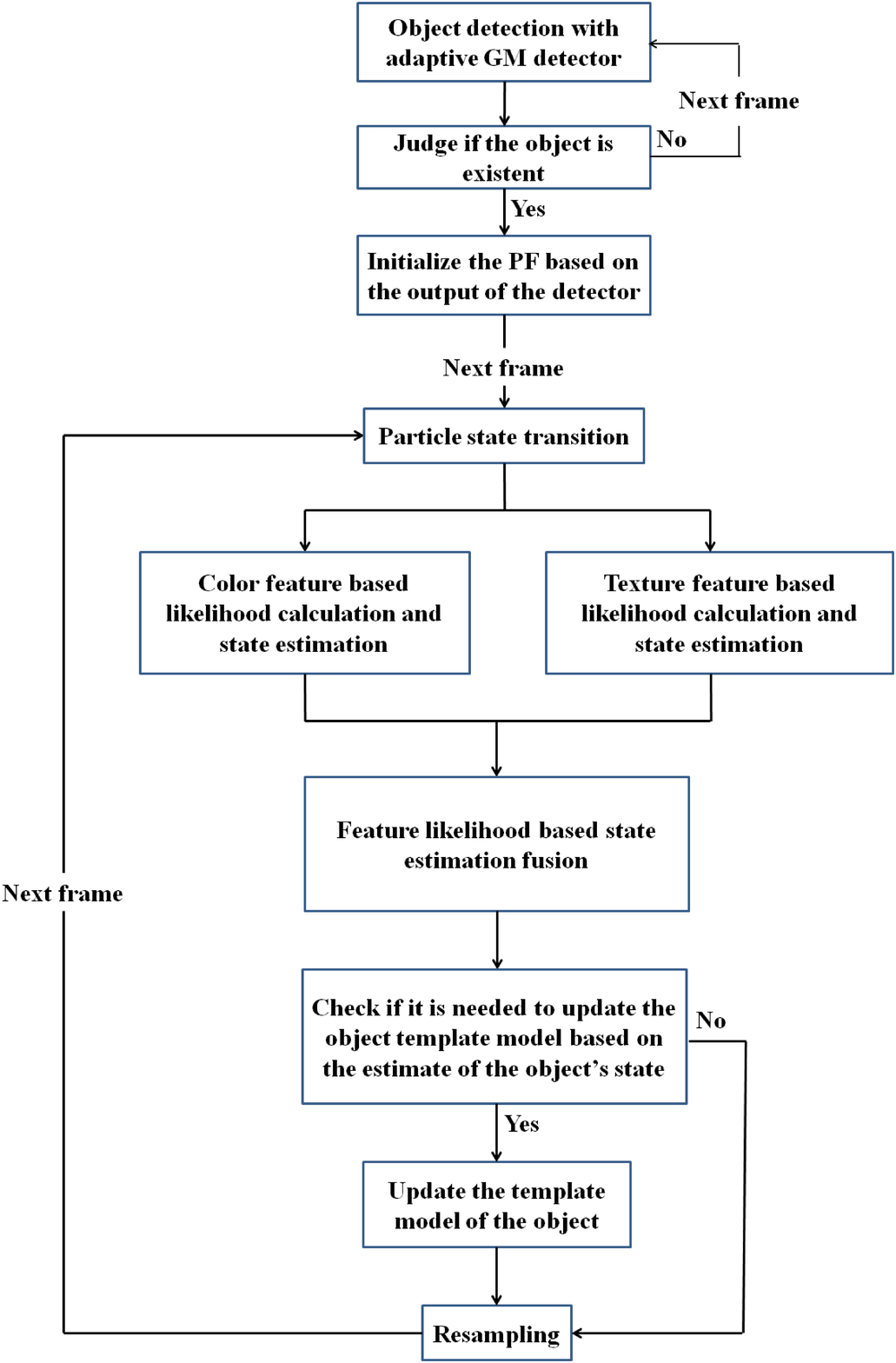}}
\end{tabular}
\caption{A working flow chart of the proposed algorithm}\label{flow}
\end{figure}

A working flow chart of the proposed algorithm is shown in Fig.\ref{flow}. In this solution, a PF tracker is triggered by an adaptive Gaussian mixture (GM) based object detector \cite{stauffer1999adaptive}, and the color and texture features are used to represent the object and calculate the likelihood of the observations.
If an object is detected, the PF tracker is initialized by drawing a set of random samples, each denoting a hypothesises on the object's state $\textbf{\mbox{S}}$.
The state vector is defined to be $\textbf{\mbox{S}}\triangleq[l_x,v_x,l_y,v_y,h_x,h_y,a]$, where $\{l_x,l_y\}$ denotes two dimensional center location of the object, $\{v_x, v_y\}$ denotes corresponding velocity component, $h_x$ and $h_y$ denote half axes of an rectangular object region, and $a$ denotes the scale of the object region compared with the whole image of a frame. Denoting $s$ a realization of $\textbf{\mbox{S}}$, the evolution of the object state over time $t$ is specified by a Markov model as follows
\begin{equation}\label{dynamic model}
s_t=s_{t-1}+w_t,
\end{equation}
where $s_t$ denotes the value of $s$ at time step $t$, and $w$ denotes a zero mean Gaussian distributed system noise.
The goal of PF is to simulate the posterior $p(s_t|y_{1:t})$, where $y$ denotes the observation and $y_{1:t}\triangleq\{y_1,y_2,\ldots,y_t\}$. This task can be decomposed into a recursively processed prediction step \cite{arulampalam2002tutorial}
\begin{equation}
p(s_t|y_{1:t-1})=\int p(s_t|s_{t-1})p(s_{t-1}|y_{1:t-1})ds_{t-1}
\end{equation}
and update step
\begin{equation}
p(s_t|y_{1:t})=\frac{p(y_t|s_t)p(s_t|y_{1:t-1})}{\int p(y_t|s_t)p(s_t|y_{1:t-1})ds_t}.
\end{equation}
Here $p(s_t|s_{t-1})$ is the state transition prior which corresponds to the evolution of object specified by Eqn.(\ref{dynamic model}), and $p(y_t|s_t)$ is the likelihood of $y_t$ given $s_t$. Applying the principle of importance sampling, the posterior $p(s_t|y_{1:t})$ can be approximated by a finite set of weighted samples $\left\{s_t^i,\omega_t^i\right\}$ as follows
\begin{equation}
p(s_t|y_{1:t})\approx\sum_{i=1}^N \omega_t^i\delta(s_t-s_t^i),
\end{equation}
provided that the sample size $N$ is big enough. $\delta(s_t-s_t^i)$ is a Dirac delta function with mass located at $s_t^i$. The importance weight $\omega_t^i$ is normalized so that $\omega_t^i=\frac{\hat{\omega}_t^i}{\sum_{j=1}^N\hat{\omega}_t^j}$, where the particle weight $\hat{\omega}_t^i$ is given by
\begin{equation}\label{SIS_weight}
\hat{\omega}_t^i=\omega_{t-1}^i\frac{p(y_t|s_t^i)p(s_t^i|s_{t-1}^i)}{q(s_t^i|s_{0:t-1}^i,y_{1:t})}, i=1,\ldots,N,
\end{equation}
where $q(s_t^i|s_{0:t-1}^i,y_{1:t})$ is a proposal distribution from which the particles are sampled. The state transition prior is used here as the proposal, i.e., $q(s_t^i|s_{0:t-1}^i,y_{1:t})=p(s_t^i|s_{t-1}^i)$, in the same spirit as of the condensation method or the bootstrap filter \cite{isard1998condensation,smith2013sequential}. Thus we have
\begin{equation}\label{SIS_weight}
\hat{\omega}_t^i=\omega_{t-1}^i p(y_t|s_t^i), i=1,\ldots,N.
\end{equation}

The way how the likelihood $p(y|s)$ is calculated depends on which feature is selected for representing the object, see details in the following subsection.

\subsection{Object feature based likelihood calculation}
The color and texture features are selected for object representation and likelihood calculation. In what follows, the subscripts $cr$ and $tx$ are used to denote the color and texture features, respectively.

Following \cite{nummiaro2003adaptive}, we consider the color distribution inside an upright elliptic region with half axes $H_x$ and $H_y$. The color histograms are produced with function $b(\mbox{\textbf{x}})$, which assigns the color at location $\mbox{\textbf{x}}$ to the corresponding bin. Here the histograms are calculated in the RGB space using $8\times8\times8$ bins. For an object state $s$, the corresponding color distribution $p_s=\{p_s^u\}_{u=1,\ldots,m}$ is calculated as follows
\begin{equation}\label{color distribution}
p_s^u=\textbf{\mbox{C}}\sum_{j=1}^{J}\textbf{\mbox{k}}\left(\frac{\left\|s_c-\mbox{\textbf{x}}_j\right\|}{H}\right)\delta(b(\mbox{\textbf{x}}_j)-u),
\end{equation}
where $m$ is the number of bins, $J$ is the number of pixels in the region of interest, $s_c$ is the centered location of the object with state $s$, $H=\sqrt{H_x^2+H_y^2}$ is used to adapt the size of the region, the normalization factor $\textbf{\mbox{C}}=\frac{1}{\sum_{j=1}^{J}\textbf{\mbox{k}}\left(\frac{\left\|s_c-\mbox{\textbf{x}}_j\right\|}{H}\right)}$
ensures that $\sum_{u=1}^{m}p_s^u=1$, and $\textbf{\mbox{k}}$ is a weighting function defined to be
$\textbf{\mbox{k}}(r)= \begin{cases} 1-r^2 & r<1\\0 &\mbox{otherwise}
\end{cases}$, which assigns smaller weights to the pixels that are further away from the region center \cite{nummiaro2003adaptive}.
Denote $q^u$ to be the color distribution of an object template, the distance between $p_s^u$ and $q^u$ is measured using the Bhattacharyya distance as follows \cite{nummiaro2003adaptive}
\begin{equation}
d_{s,cr}=\sqrt{1-\rho(p_s^u,q^u)}),
\end{equation}
where $\rho(p_s^u,q^u)=\sum_{u=1}^m\sqrt{p_s^uq^u}$.
The color feature based likelihood is calculated as follows
\begin{equation}
p_{cr}(y|s)=\frac{1}{\sqrt{2\pi}\sigma_{cr}}\exp\left(-\frac{d_{s,cr}^2}{2\sigma_{cr}^2}\right),
\end{equation}
where $\sigma_{cr}$ is set to be 0.1. In the similar way, we can calculate the texture feature based likelihood as below
\begin{equation}
p_{tx}(y|s)=\frac{1}{\sqrt{2\pi}\sigma_{tx}}\exp\left(-\frac{d_{s,tx}^2}{2\sigma_{tx}^2}\right),
\end{equation}
where $\sigma_{tx}=0.1$, $d_{s,tx}$ is obtained in the similar way as for $d_{s,cr}$, while the only difference lies in how to derive the corresponding $p_s^u$ and $q^u$. The local binary pattern (LBP) technique is used for extracting the texture feature. See details on the LBP operator in \cite{ahonen2006face} and refer \cite{ning2009robust} for LBP based derivation of $p_s^u$ and $q^u$.
\subsection{Feature likelihood based state estimation fusion}
Substituting $p(y_t|s_t^i)$ in Eqn. (\ref{SIS_weight}) with $p_{cr}(y_t|s_t^i)$ and $p_{tx}(y_t|s_t^i)$ derived in the above subsection, we respectively obtain importance weights $\hat{\omega}_{t,cr}^i$ and $\hat{\omega}_{t,tx}^i$. Then the normalized weights $\omega_{t,cr}^i$ and $\omega_{t,tx}^i$ are available. Thus the object state estimates based on the color and texture features are just $\hat{s}_{t,cr}=\sum_{i=1}^N\omega_{t,cr}^is_t^i$ and $\hat{s}_{t,tx}=\sum_{i=1}^N\omega_{t,tx}^is_t^i$, respectively.

We propose here a marginal feature likelihood based state fusion approach. Take the $t$th iteration for example, just after the particle state transition procedure, we will have a weighted sample set $\{\omega_{t-1}^i,s_t^i\}_{i=1}^N$ at hand.
Then the marginal likelihood of the color feature is shown to be
\begin{equation}\label{marginal_lik_color}
L_{t,cr}=\sum_{i=1}^N\omega_{t-1}^ip_{cr}(y_t|s_t^i).
\end{equation}
The marginal likelihood of the texture feature $L_{t,tx}$ is calculated in the same way.
Using a uniform prior on the color and texture features, we get the posterior probabilities of the color and texture features, namely $\mbox{Pr}_{t,cr}=\frac{L_{t,cr}}{L_{t,cr}+L_{t,tx}}$ and $\mbox{Pr}_{t,tx}=\frac{L_{t,tx}}{L_{t,cr}+L_{t,tx}}$.
Then we do feature fusion to yield the finally outputted state estimate as follows
\begin{equation}
\hat{s}_t=\mbox{Pr}_{t,cr}\hat{s}_{t,cr}+\mbox{Pr}_{t,tx}\hat{s}_{t,tx}.
\end{equation}
The corresponding normalized importance weight of $s_t^i$ is calculated as below
\begin{equation}
\omega_t^i=\mbox{Pr}_{t,cr}\omega_{t,cr}^i+\mbox{Pr}_{t,tx}\omega_{t,tx}^i, i=1,\ldots,N,
 \end{equation}
 based on which the resampling step is ran to prevent from particle divergence \cite{arulampalam2002tutorial}.
\subsection{Update the object template model online}
In the proposed algorithm, see Fig.\ref{flow}, the initial object template model is built based on the output of the GM detector.
Once the detector reports a detection, the color and texture features based templates, represented by the color and LBP histograms, will be calculated. Afterwards, the marginal likelihoods of the color and texture features will be calculated for each frame signal, see Eqn.(\ref{marginal_lik_color}). At the $t$th ($t>5$) iteration, first calculate the average and the corresponding standard error of the marginal likelihoods in the previous 5 iterations. If the absolute Euclidean distance between the marginal likelihood of current iteration and the averaged value is further than the double of the standard error, we determine that an abrupt change in the distribution of this feature happened. If the distributions of both features changed abruptly, we determine that an occlusion has happened. In this case, we use a one-step prediction as the estimate of the object state and do not need to update any template. If only one feature's distribution changed abruptly, we determine that its template model needs to be updated and then update it to be the distribution of current iteration.
\section{Experimental Results}
To evaluate the performance, we tested our algorithm in detecting and tracking dim moving object in two pieces of complex real life video signals. The existing alternatives, including ALG I: PF using adaptive color feature \cite{nummiaro2003adaptive}, ALG II: the PF using LBP modeled texture feature \cite{ye2010face}, ALG III: the PF using equally weighted texture and color features \cite{ying2010particle} and ALG IV: the adaptive GM detector \cite{stauffer1999adaptive}, are involved for comparison.
\subsection{Case I}\label{sec:case1}
The first video is taken in a dark tunnel, and the task is to detect and then track a moving car which is passing through this tunnel. The object appears in the 38th frame, and its color is similar with the road in the video background. In the last frames, the taillights of the car light up leading to a change in the color feature distribution of the object.

\begin{figure}[htb]
\begin{tabular}{c}
\centerline{\includegraphics[width=1.25in,height=1in]{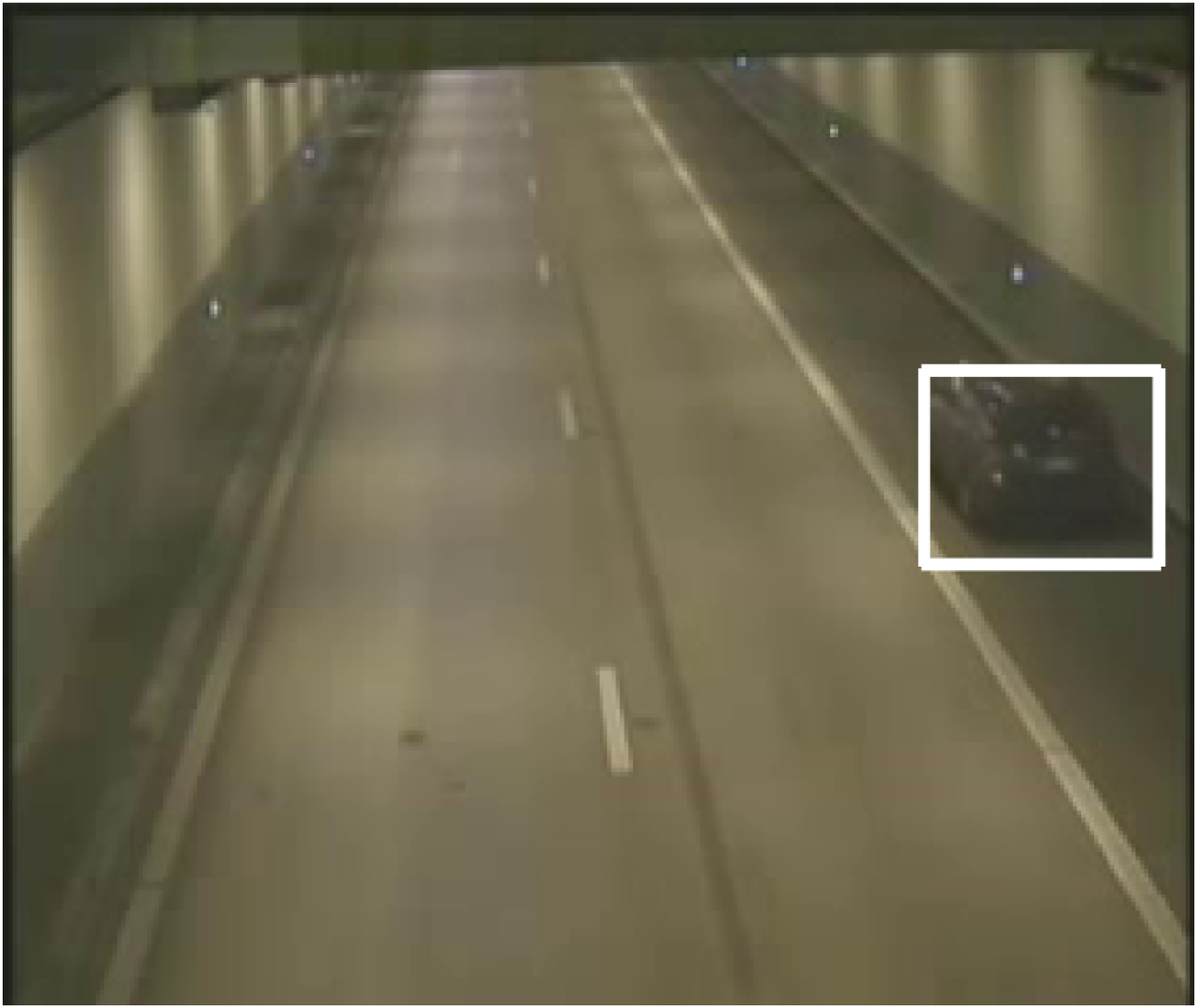}\includegraphics[width=1.25in,height=1in]{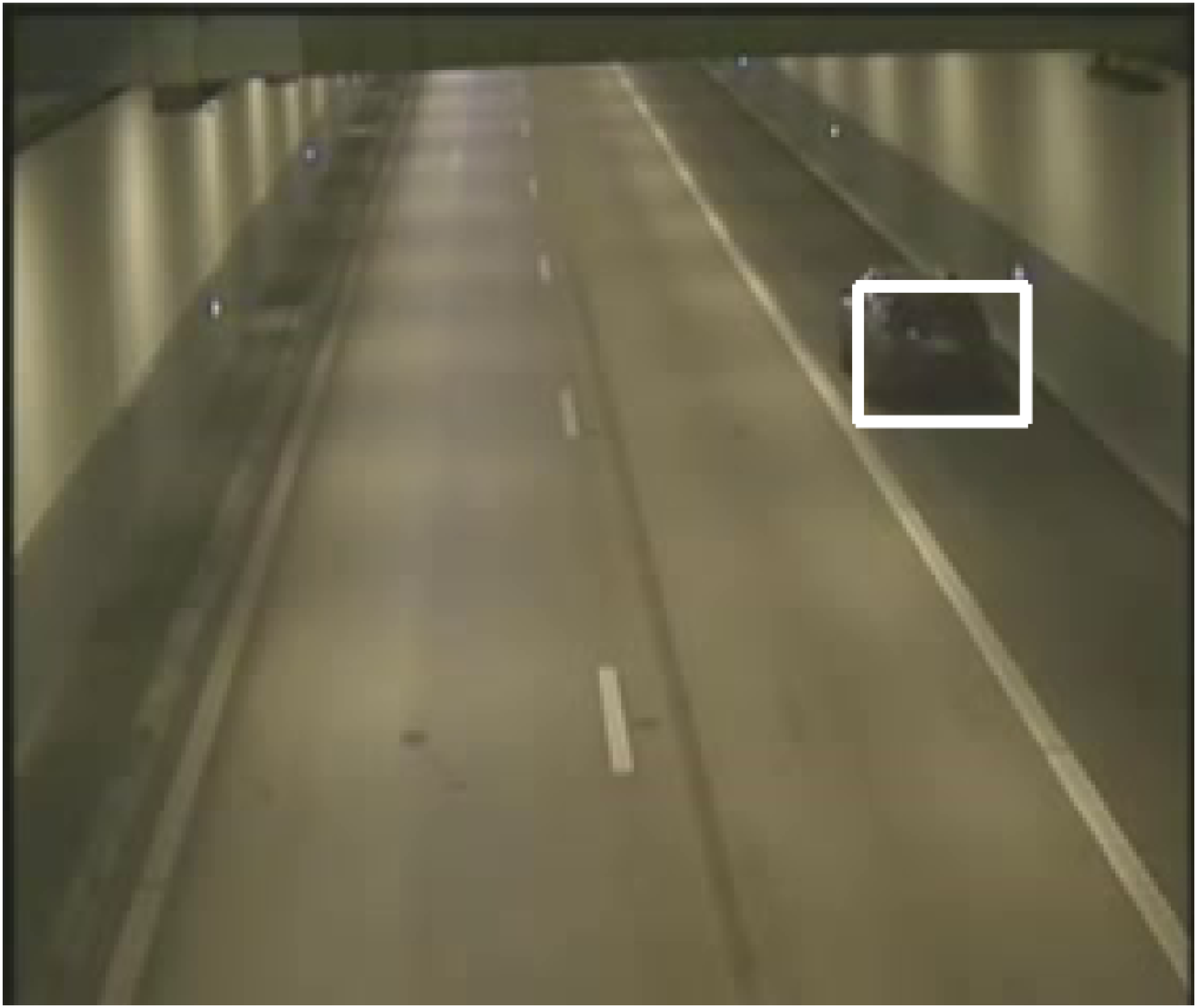}}\\
\centerline{\includegraphics[width=1.25in,height=1in]{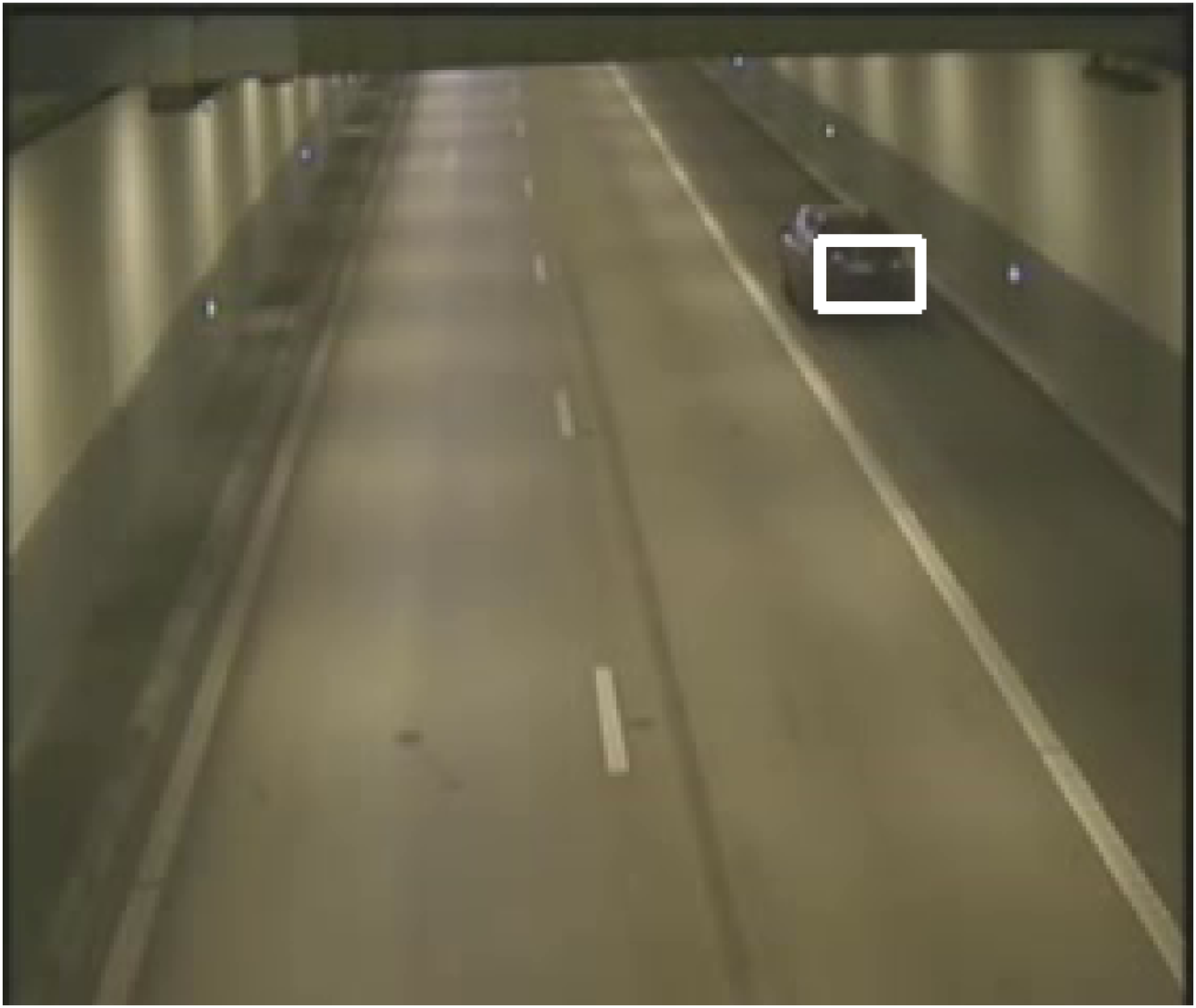}\includegraphics[width=1.25in,height=1in]{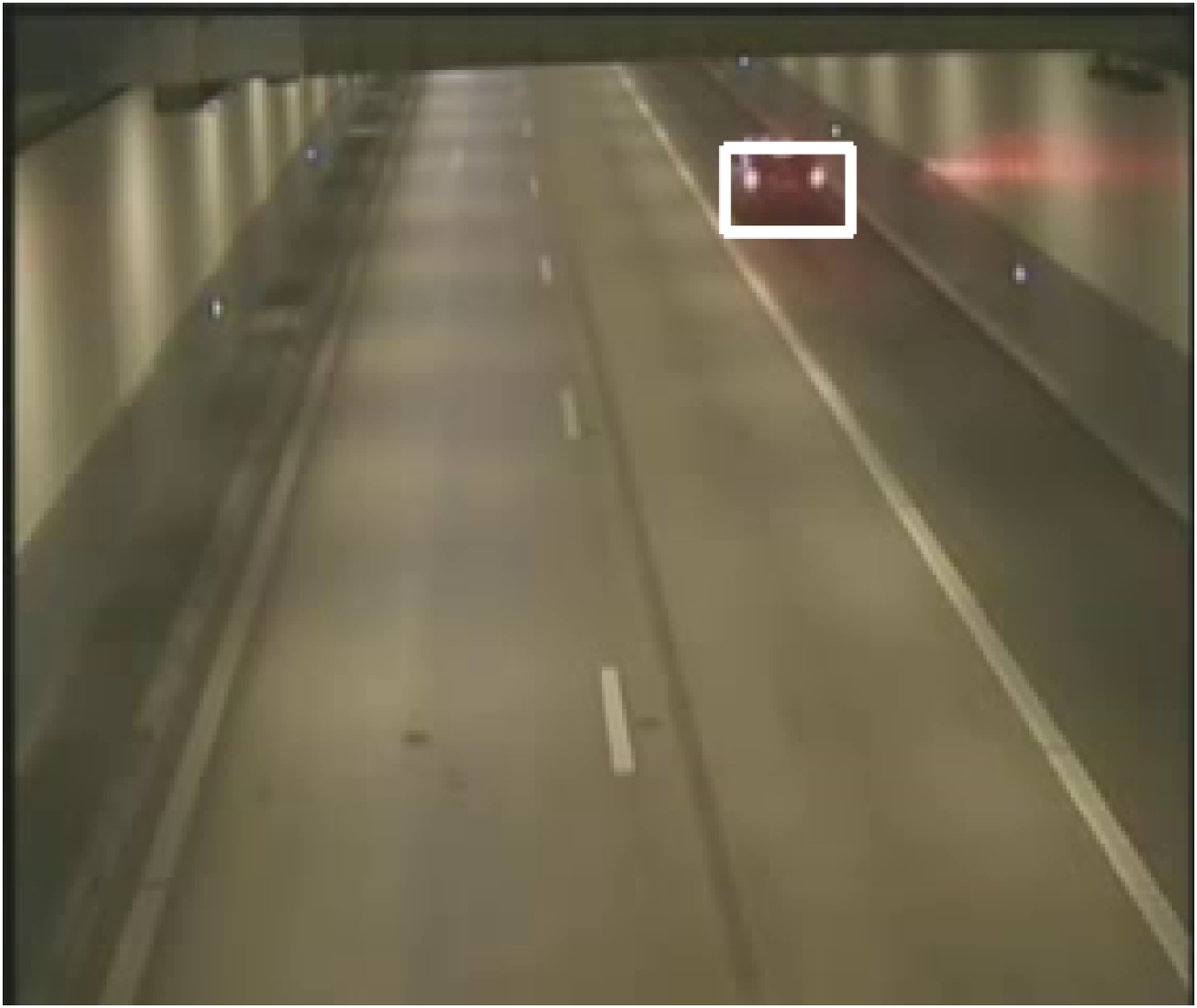}}
\end{tabular}
\caption{Tracking results of the proposed algorithm for case I} \label{case1_track}
\end{figure}

\begin{figure}[htb]
\begin{tabular}{c}
\centerline{\includegraphics[width=3.5in,height=2.0in]{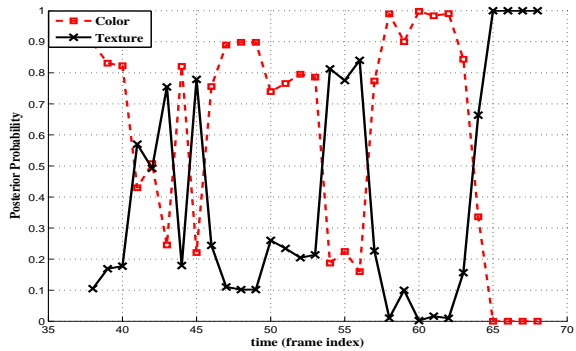}}
\end{tabular}
\caption{Features' posterior probabilities in test case I.} \label{prob_feature_caseI}
\end{figure}

The result of an example run of the proposed algorithm is shown in Fig.\ref{case1_track}. As is shown, it is not influenced by the presence of confusing colors in the background, abrupt changes in the object's color feature and scale changes of the object. Indicated by the change in the size of the white box that corresponds to the detected object region, we can infer that the object template model is updated successfully online. The changes in the posterior probabilities of the color and texture features are examined and illustrated in Fig.\ref{prob_feature_caseI}. It is shown that the feature fusion effect is truly taking effect in dynamically adjusting the usages of the color and texture features in the tracking process.

\begin{figure}[htb]
\begin{tabular}{c}
\centerline{\includegraphics[width=3in,height=2in]{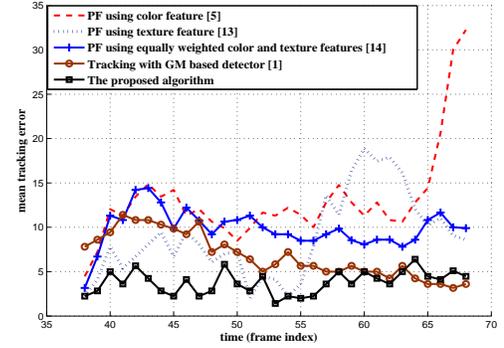}}
\end{tabular}
\caption{Mean tracking error in test case I. The size of a whole image in one frame is 320$\times$240.} \label{error_caseI}
\end{figure}

A numerical performance comparison resulted from a 100 times independent runs of each involved algorithm is shown in Fig.\ref{error_caseI}. It is shown that the proposed algorithm performs best, while performance of ALG I \cite{nummiaro2003adaptive} gets deteriorated remarkably once the taillights light up. The computation time required to run each algorithm, based on a fixed computing environment, is presented in Table 1.

Table $1$: Computation time comparison (unit: second)

\begin{tabular}{c|c|c|c|c}\centering\small
       ALG I &ALG II&ALG III&ALG IV&the proposed\\
\hline 3.404&5.668&7.797&9.255&8.632\\
\hline
\end{tabular}
\subsection{Case II}\label{sec:case2}
The second task is to track a radio controlled and green colored helicopter model, which flies up and down. In this video, the helicopter model first appears in the sky, flies downwards and get partially occluded by the trees and a human operator, and then rises up again. Then it falls downwards again, with its body mixed up with the trees behind it.

\begin{figure}[htb]
\begin{tabular}{c}
\centerline{\includegraphics[width=1.25in,height=1in]{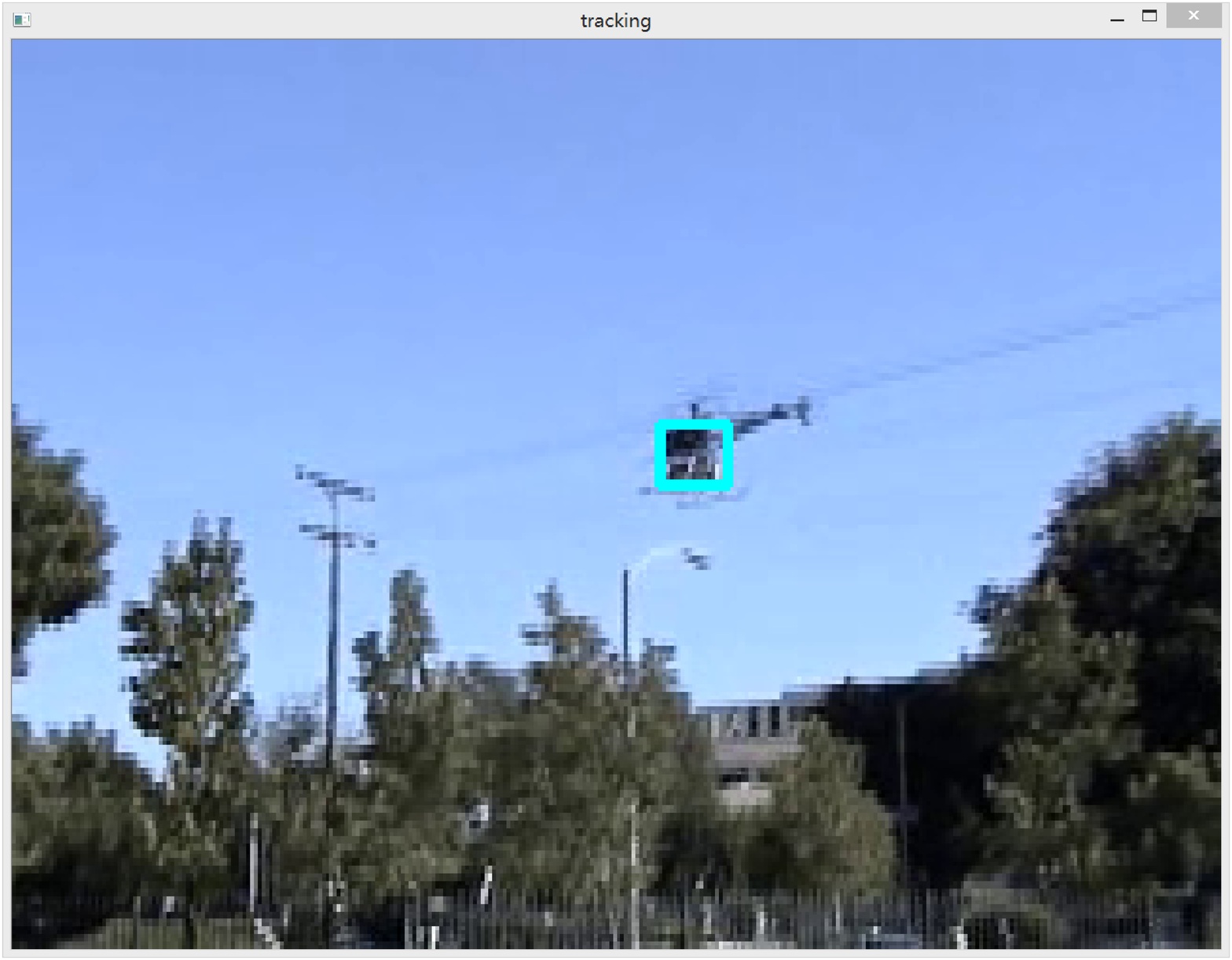}\includegraphics[width=1.25in,height=1in]{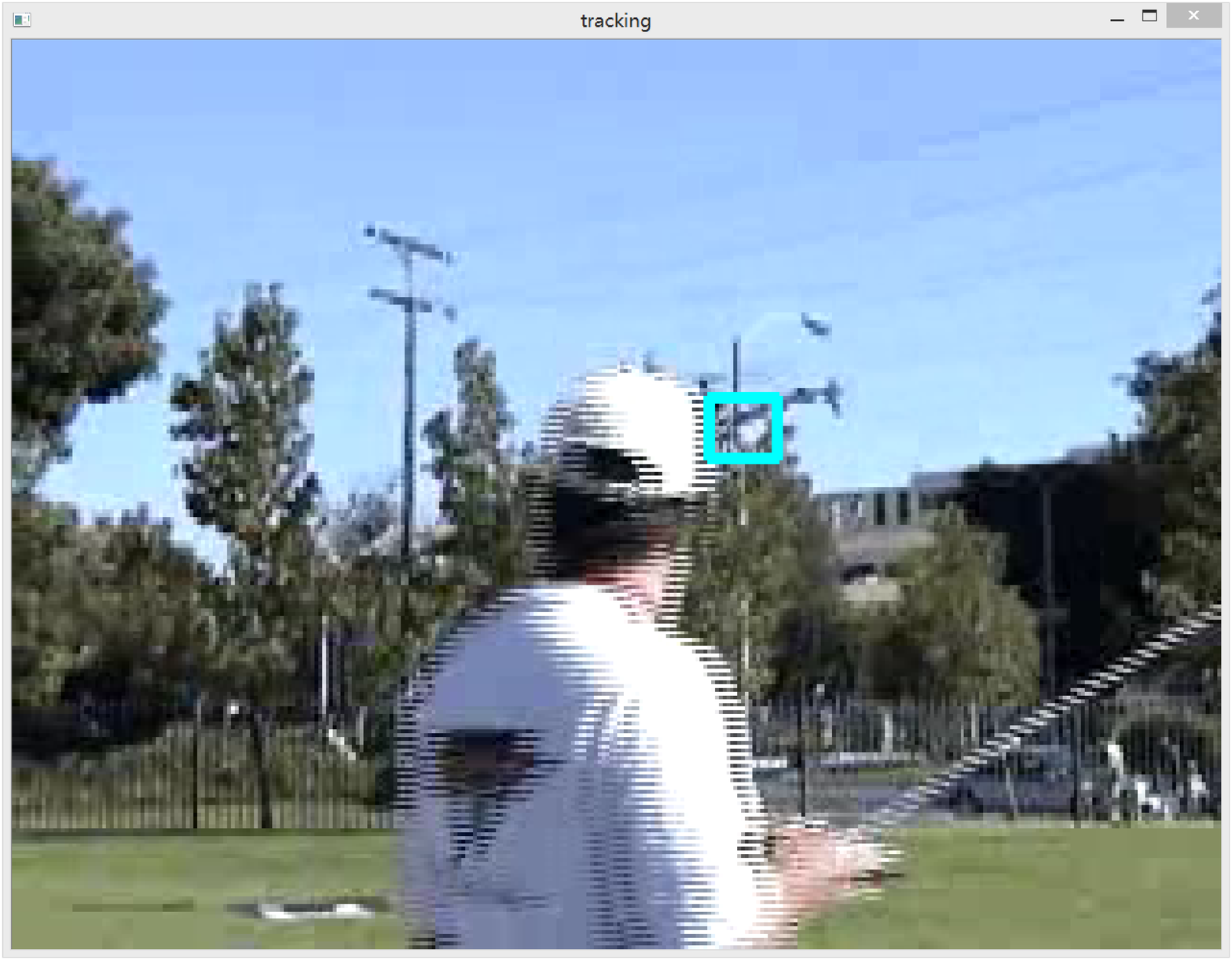}}\\
\centerline{\includegraphics[width=1.25in,height=1in]{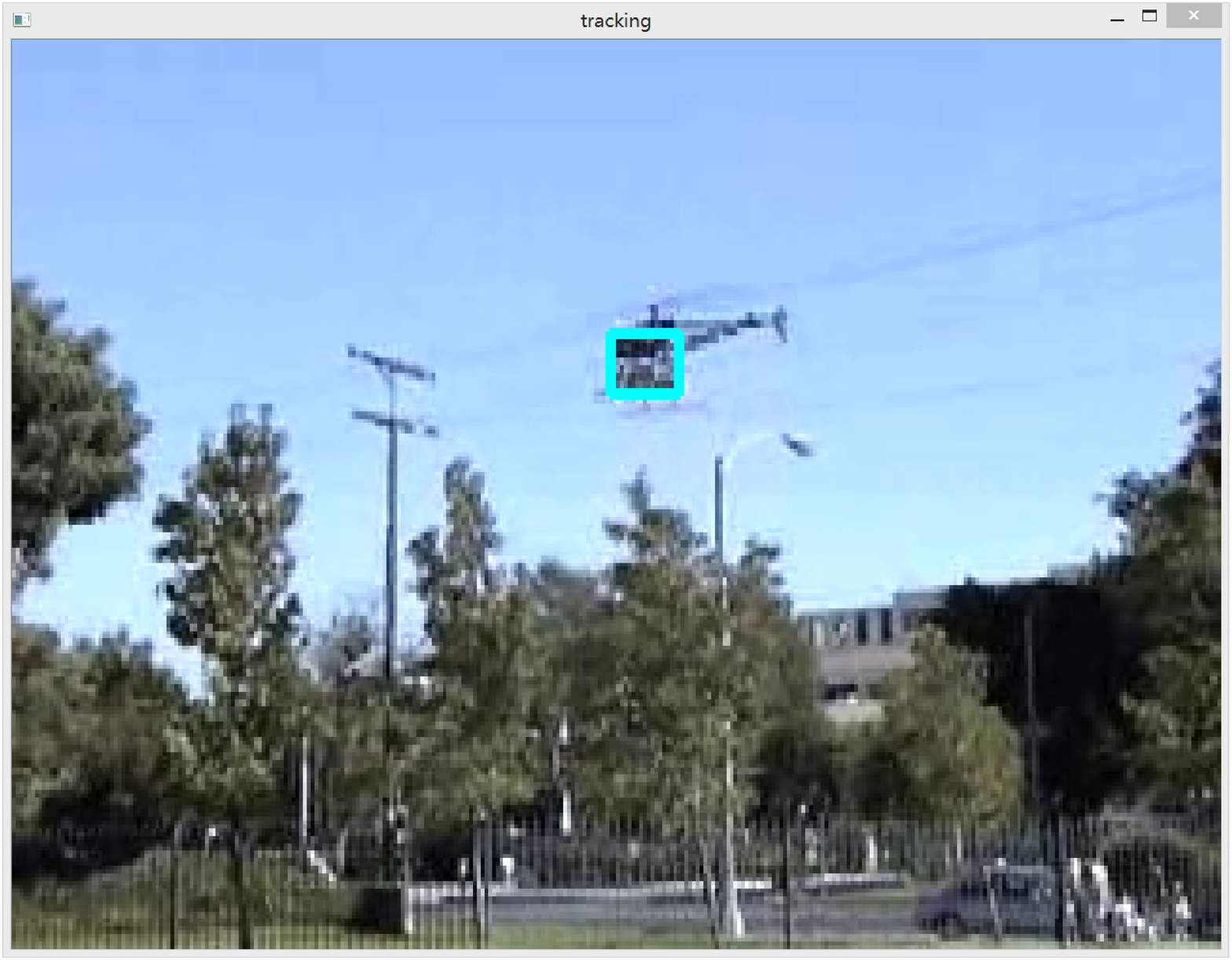}\includegraphics[width=1.25in,height=1in]{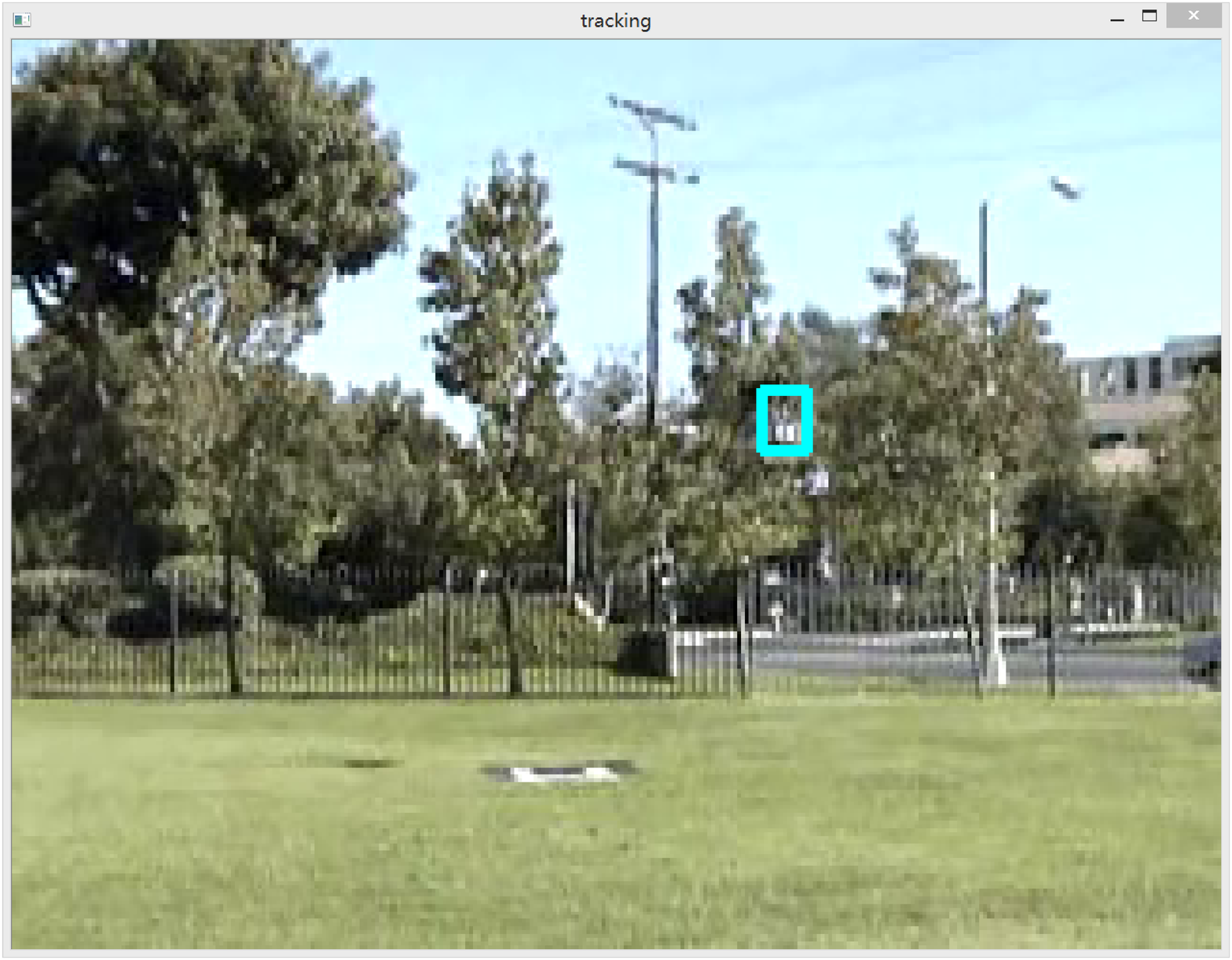}}
\end{tabular}
\caption{Tracking results of the proposed algorithm for case II} \label{case2_track}
\end{figure}

In this case, the proposed algorithm again works very well in tracking the object accurately from beginning to end. See Fig.\ref{case2_track} for 4 typical frames. ALG I \cite{nummiaro2003adaptive} fails when the body of the object is totally mixed up with the trees, as it erroneously identifies the crown of a tree as the object. ALG III \cite{ying2010particle} performs similarly as ALG I. An early failure happens to ALG II \cite{ye2010face} when the object approaches the trees in the first time.

\section{Conclusion}
A PF tracking algorithm is proposed based on Gaussian mixture background modeling, color and texture features fusion and adaptive object template updating, see Fig.\ref{flow} for a working flow chart. This method is shown to be robust against partial occlusion (Sec.\ref{sec:case2}), presence of confusing colors in the background (Sec.\ref{sec:case1} and \ref{sec:case2}), abrupt changes in the object's feature (Sec.\ref{sec:case1}), scale changes of the object (Sec.\ref{sec:case1}). The advantages of this method over existing alternatives stem from a new theoretically sound feature fusion mechanism, the corresponding online approach to update the object template model and an original mixture of the efficient algorithmic components. The feature fusion approach presented here can be adapted to do fusions of more features, which may be helpful in dealing with more complex scenarios. Extending the work here to track multiple video objects is also possible.

\bibliographystyle{IEEEbib}
\bibliography{mybibfile}

\begin{thebibliography}{10}

\bibitem{stauffer1999adaptive}
C.~Stauffer and W.~Grimson,
\newblock ``Adaptive background mixture models for real-time tracking,''
\newblock in {\em Proc. of IEEE Conf. on Computer Vision and Pattern
  Recognition (CVPR)}, 1999, vol.~2, pp. 246--252.

\bibitem{arulampalam2002tutorial}
M.~S. Arulampalam, S.~Maskell, N.~Gordon, and T.~Clapp,
\newblock ``A tutorial on particle filters for online nonlinear/non-gaussian
  bayesian tracking,''
\newblock {\em IEEE Trans. on Signal Processing}, vol. 50, no. 2, pp. 174--188,
  2002.

\bibitem{doucet2000sequential}
A.~Doucet, S.~Godsill, and C.~Andrieu,
\newblock ``On sequential monte carlo sampling methods for bayesian
  filtering,''
\newblock {\em Statistics and computing}, vol. 10, no. 3, pp. 197--208, 2000.

\bibitem{takala2007multi}
V.~Takala and M.~Pietikainen,
\newblock ``Multi-object tracking using color, texture and motion,''
\newblock in {\em Proc. of IEEE Conf. on Computer Vision and Pattern
  Recognition (CVPR)}, 2007, pp. 1--7.

\bibitem{nummiaro2003adaptive}
Katja Nummiaro, Esther Koller-Meier, and Luc Van~Gool,
\newblock ``An adaptive color-based particle filter,''
\newblock {\em Image and vision computing}, vol. 21, no. 1, pp. 99--110, 2003.

\bibitem{yang2005fast}
C.~Yang, R.~Duraiswami, and L.~Davis,
\newblock ``Fast multiple object tracking via a hierarchical particle filter,''
\newblock in {\em Proc. of 10th IEEE Intl. Conf. on Computer Vision (ICCV)},
  2005, vol.~1, pp. 212--219.

\bibitem{vacchetti2004combining}
L.~Vacchetti, V.~Lepetit, and P.~Fua,
\newblock ``Combining edge and texture information for real-time accurate 3{D}
  camera tracking,''
\newblock in {\em Proc. of 3rd IEEE and ACM Intl. Symp. on Mixed and Augmented
  Reality (ISMAR)}, 2004, pp. 48--56.

\bibitem{yilmaz2006object}
A.~Yilmaz, O.~Javed, and M.~Shah,
\newblock ``Object tracking: A survey,''
\newblock {\em ACM Computing Surveys}, vol. 38, no. 4, pp. 13, 2006.

\bibitem{isard1998condensation}
M.~Isard and A.~Blake,
\newblock ``Condensation—conditional density propagation for visual
  tracking,''
\newblock {\em International journal of computer vision}, vol. 29, no. 1, pp.
  5--28, 1998.

\bibitem{smith2013sequential}
A.~Smith, A.~Doucet, N.~de~Freitas, and N.~Gordon,
\newblock {\em Sequential Monte Carlo methods in practice},
\newblock Springer Science \& Business Media, 2013.

\bibitem{ahonen2006face}
T.~Ahonen, A.~Hadid, and M.~Pietikainen,
\newblock ``Face description with local binary patterns: Application to face
  recognition,''
\newblock {\em IEEE Trans. on Pattern Analysis and Machine Intelligence}, vol.
  28, no. 12, pp. 2037--2041, 2006.

\bibitem{ning2009robust}
J.~Ning, L.~Zhang, D.~Zhang, and C.~Wu,
\newblock ``Robust object tracking using joint color-texture histogram,''
\newblock {\em International Journal of Pattern Recognition and Artificial
  Intelligence}, vol. 23, no. 07, pp. 1245--1263, 2009.

\bibitem{ye2010face}
J.~Ye, Z.~Liu, and J.~Zhang,
\newblock ``A face tracking algorithm based on {LBP} histograms and particle
  filtering,''
\newblock in {\em Proc. of 6th Intl. Conf. on Natural Computation (ICNC)},
  2010, vol.~7, pp. 3550--3553.

\bibitem{ying2010particle}
H.~Ying, X.~Qiu, J.~Song, and X.~Ren,
\newblock ``Particle filtering object tracking based on texture and color,''
\newblock in {\em Proc. of Intl. Symp. on Intelligence Information Processing
  and Trusted Computing (IPTC)}, 2010, pp. 626--630.

\end{thebibliography}
\end{document}